\documentclass[letterpaper, 10 pt, conference]{ieeeconf}  %

\IEEEoverridecommandlockouts                              %

\let\labelindent\relax
\usepackage{graphics,enumitem} %
\usepackage{epsfig} %
\usepackage{mathptmx} %
\usepackage{times} %
\usepackage{amsmath} %
\usepackage{amssymb}  %
\usepackage{booktabs}
\usepackage{multirow}
\usepackage{algpseudocode}
\usepackage{algorithm}
\usepackage{makecell}
\usepackage{caption}
\usepackage{subcaption}
\usepackage{url}
\usepackage{multicol}

\usepackage{cite}
\usepackage{balance}
\usepackage[pagebackref=true,breaklinks=true,colorlinks=true,bookmarks=false,citecolor=blue]{hyperref}
\hypersetup{
colorlinks=true,
linkcolor=blue,
filecolor=magenta,      
urlcolor=blue,
citecolor=blue
}
\newcommand\mypara[1]{\vspace{5pt}\noindent\textbf{#1}$\quad$}

\title{\LARGE \bf Legs as Manipulator: Pushing Quadrupedal Agility Beyond Locomotion}

\author{Xuxin Cheng\\
Carnegie Mellon University \and
Ashish Kumar \\
UC Berkeley \and
Deepak Pathak \\
Carnegie Mellon University
}

\begin{document}
\maketitle
\begin{abstract}
Locomotion has seen dramatic progress for walking or running across challenging terrains. However, robotic quadrupeds are still far behind their biological counterparts, such as dogs, which display a variety of agile skills and can use the legs beyond locomotion to perform several basic manipulation tasks like interacting with objects and climbing. In this paper, we take a step towards bridging this gap by training quadruped robots not only to walk but also to use the front legs to climb walls, press buttons, and perform object interaction in the real world. To handle this challenging optimization, we decouple the skill learning broadly into \textit{locomotion}, which involves anything that involves movement whether via walking or climbing a wall, and \textit{manipulation}, which involves using one leg to interact while balancing on the other three legs. These skills are trained in simulation using curriculum and transferred to the real world using our proposed sim2real variant that builds upon recent locomotion success. Finally, we combine these skills into a \textit{robust} long-term plan by learning a behavior tree that encodes a high-level task hierarchy from one clean expert demonstration. We evaluate our method in both simulation and real-world showing successful executions of both short as well as long-range tasks and how robustness helps confront external perturbations. Videos at~\url{https://robot-skills.github.io}
\end{abstract}

\vspace{-0.05in}
\section{Introduction}
\vspace{-0.05in}
Robotic quadrupeds and bipeds can walk across challenging scenarios ranging from hiking on hills to walking over rocky surfaces near river beds~\cite{Miki2022-ez,kumar2021rma,fu2021minimizing,RoboImitationPeng20,CassieLi2021,rudin2022learning,Ji2022-lt,agarwal2022legged,yang2022fast,Margolis2022-ju,li2023robust}. What is next for legged systems? Despite being highly effective walkers, legged robots are far behind the dexterity and agility of quadrupeds in the animal kingdom, such as dogs or cats, who can use their legs more generally than just for the task of walking. In particular, quadrupeds sometimes use their legs to open a door, dig a hole, pull an object, etc.~\cite{pets18}. Indeed, locomotion and manipulation can be seen as dual of each other~\cite{johnson2013legged,lynch1996nonprehensile} and share evolutionary origins where frontal legs evolved to become arms in bipeds. Providing robotic quadrupeds a similar ability would not only push the \textit{agility} of robotic locomotion but can also greatly expand the reach as well as functionality even if there is an arm already attached on top of the quadruped.

\begin{figure}[t]
\centering
\includegraphics[width=1.0\linewidth]{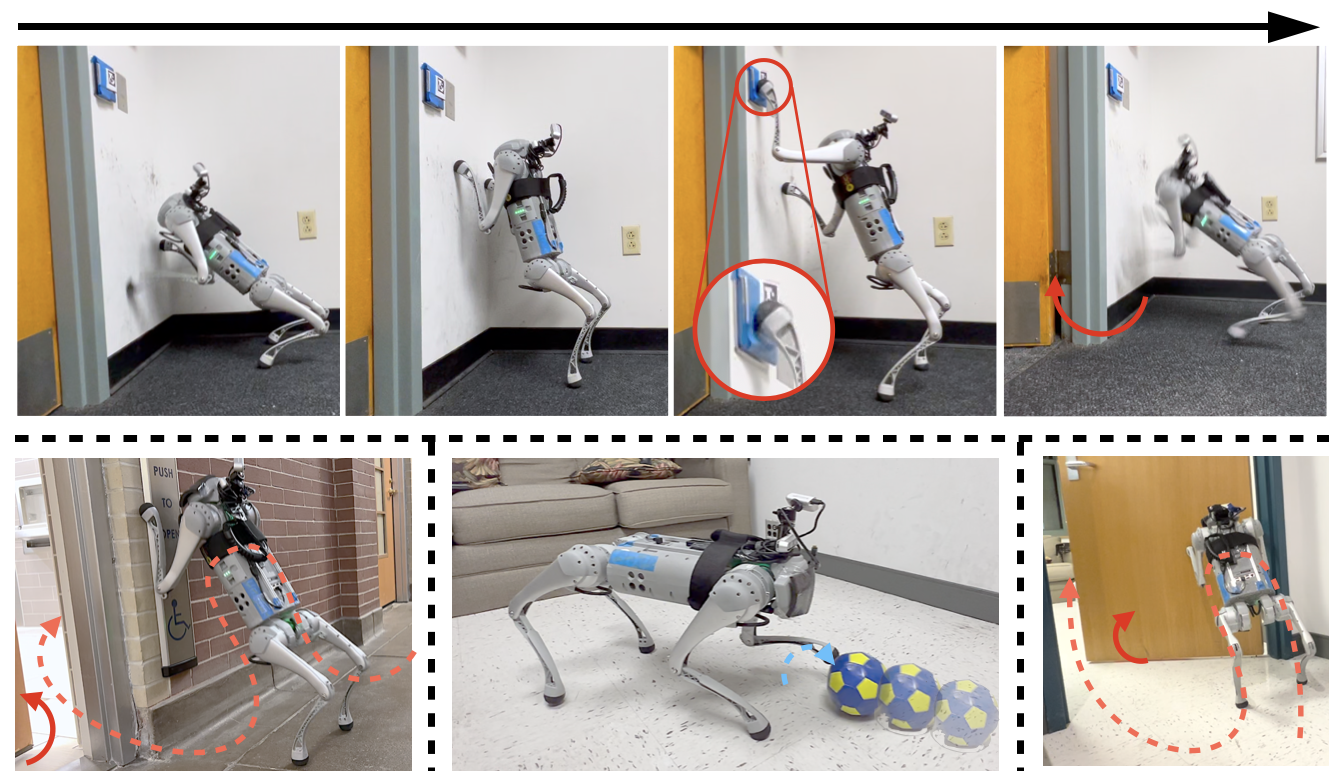}
\vspace{-0.2in}
\caption{\small Examples of real-world skills. Top row: Robot climbs high on the wall to operate a wheelchair access button using its leg and then get off the wall to walk out of the door. Bottom left: Robot climbs on the wall and uses its weight to press the button. Bottom middle: Robot kicks a ball on the ground. Bottom right: Robot climbs a door and opens it using its weight and then walk indoors.
}
\label{fig:teaser}
\vspace{-0.2in}
\end{figure}

In this work, we focus on this joint problem of learning locomotion as well as basic manipulation skills by expanding the capability of quadruped robots to enable them to use their legs as manipulators. In particular, as shown in Fig.~\ref{fig:teaser}, we focus on tasks like climbing a wall with front legs, jumping on a wall to reach a button, using a leg to push a button, etc; and then combining them to achieve long-range behaviors. More broadly, we follow the popular approach of learning environment latent conditioned policies in simulation using reinforcement learning (RL) and then transferring them to the real world via sim2real~\cite{Lee2020-my, kumar2021rma}.

A major challenge in this technique for training legs to simultaneously walk and perform skills like climbing is that the RL can get stuck in local minima. To get around this issue, we decouple the skill policy training into \textit{locomotion} and \textit{manipulation} using legs. Locomotion policy captures anything that involves movement such as walking or climbing onto a wall and is conditioned on the commanded linear as well as angular velocity commands. On the other hand, the manipulation policy captures moving a single leg for tasks like pressing the button using one foot while balancing on other legs. This button pressing can be performed on a horizontal ground or a vertical wall.

\begin{figure*}[t!]
    \centering
    \includegraphics[width=.98\linewidth]{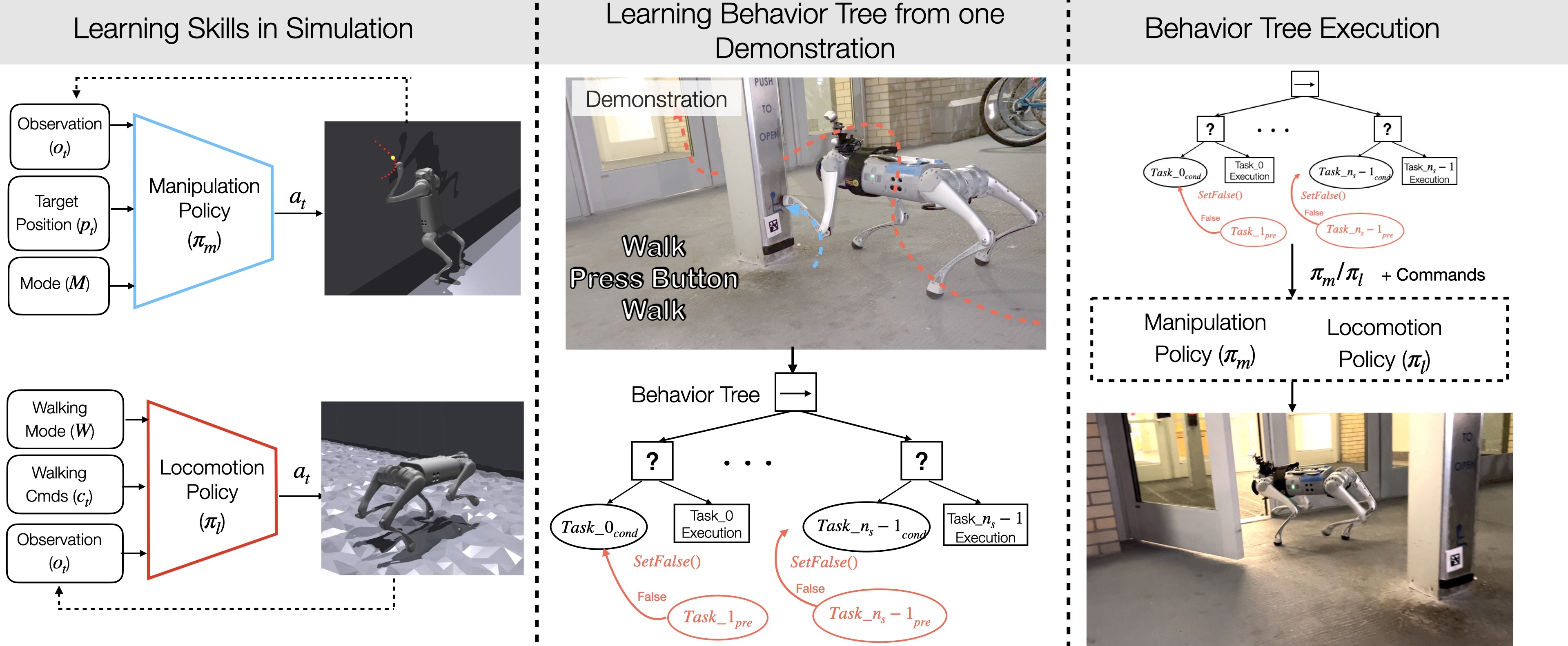}
    \vspace{-0.08in}
    \caption{\small Overview of the method. Left: Skill learning framework where the two policies are trained separately with their own observations $o_t^l$ and $o_t^m$ in simulation. The user commands include boolean values to control different modes of walking (walk vs climb), to activate button pressing, and velocity commands. Middle: Learning a behavior tree from a single high-level demonstration which is primarily responsible for selecting between the walking and manipulating policies, as well as the mode to activate. Right: Once the behavior tree is learned, we execute the behavior tree under external interruptions and perturbations to show robust execution. The $\longrightarrow{}$ represents a Sequence node that executes its children in left-to-right order until all children return success. The $?$ is a Selector node that returns success or running whichever it encounters first in its children. The oval is a condition node that returns success when the condition is met and failure otherwise. The rectangle is the execution node that always return running. Videos at~\url{https://robot-skills.github.io/}.}
    \label{fig:method} 
    \vspace{-0.15in}
\end{figure*}

We train these skills in simulation and transfer them to the real world. However, skills like climbing are highly agile and diverse in their behavior, thus relying on just onboard sensors is not sufficient. To stabilize the training, we need additional state estimates (velocity and foot contacts) which may not be directly available via a sensor. To resolve this, we follow an approach similar to RMA~\cite{kumar2021rma} to get an estimate of a unified state estimator (USE) via proprioception history, which can then be reliably computed onboard. Once we have policies that work in simulation, we use our proposed variant of regularized online adaptation (ROA)~\cite{Fu2022-si} to achieve a higher performance than off-the-shelf methods such as RMA~\cite{kumar2021rma}.

Once we have a quadruped that can perform diverse skills, it must also figure out how to stitch these skills together to perform long-term planning tasks in a manner that is robust to the failures in underlying skills~\cite{pertsch2022star, Peng2021-sq, Peng2022-ez}. We propose to leverage the classic idea of behavior trees~\cite{ogren2018bt,bates1992integrating} and adapt it to our learning paradigm. In particular, after training low-level skills, we learn a behavior tree from a single clean demonstration of a long-range task and show a robust replay of the behavior despite interruptions by introducing visually matched preconditions, recording the success of a low-level skill on completion, and a recursive backing up behavior under unexpected interruptions and perturbations. 

We show extensive evaluation in both simulation and the real world. We use the learned skills to perform a variety of real-world tasks where robot climbs and presses buttons on walls or the ground. 

\vspace{-0.05in}
\section{Related Work}
Many works study locomotion and some works study using legs for manipulation. However, few works have truly investigated how to combine locomotion and manipulation skills for real-world tasks. We review these below.
\subsubsection{Legged Locomotion}
Recent advances in reinforcement learning have enabled a set of works that train a neural network based walking controller for legged robots \cite{CassieLi2021, kumar2021rma, Miki2022-ez, fu2021minimizing, RoboImitationPeng20, rudin2022learning, Ji2022-lt,agarwal2022legged}. How to achieve various gait patterns for quadrupedal robots that are emergent on real dogs is studied in \cite{yang2022fast, fu2021minimizing, Margolis2022-ju}. \cite{Margolis2022-ju, CassieLi2021} train gait parameters conditioned policies to get diverse walking behavior that can achieve simple object manipulation such as pitching down to unload a ball on the back of the robot. However, these emergent individual skills are not dexterous enough and have not been synthesized to accomplish more complex tasks.
\subsubsection{Legged Manipulation}
Manipulation via locomotion has been studied in \cite{nachum2019multi, yang2020dynamic, Shi2020-en, Ding2016-xx, Ji2022-wy}. \cite{Shi2020-en} proposed to use all the four legs of a quadrupedal robot as a dexterous manipulator by lying the robot back on the ground. However this scarifies the mobility of the legged system and makes it no difference from a four finger manipulator. \cite{Ji2022-wy} enables the quadruped to kick a ball to a goal with real-world finetuning but is still a single skill result without any agility in locomotion and lacks generalization to long-horizon tasks.
\subsubsection{Skill Synthesizing}
The ability to synthesize a vast variety of motor skills is important to achieve long-horizon complex behaviors for robots. Prior works \cite{Peng2021-sq, Peng2022-ez} combine adversarial imitation learning and unsupervised reinforcement learning to develop skill embeddings, and then train a high level policy to synthesize these low-level skill conditioned on task-specific rewards.
\cite{lee2021adversarial, lee2018composing} aim to address the problem of transitions between different skills to chain long-horizon tasks whose success rate can be exponentially reducing due to transition failures.
\subsubsection{Behavior Trees}
Long-horizon tasks can be expressed with various forms of abstractions including decision trees, finite state machines or a neural network. Behavior trees are one of these forms that are firstly vastly used in game industry for non-player character control. 
With its modularity, transparency and execution speed, they are starting to be used more in robotics \cite{French2019-ah, Marzinotto2014-hi, ogren2018bt, bates1992integrating}.
The modularity helps the user to modify the behaviors without affecting other parts. The transparency gives more ease understanding the behaviors. And execution speed ensures reactive behaviors encountered with unknown environment changes.

\section{Method: Legs as Manipulators}
Our proposed skill learning and composition framework is shown in Fig.~\ref{fig:method}. We first learn low level skills in simulation and transfer them to the real world via online adaptation. Once we have these skills available in the real world, we then use behavior trees to learn to compose these skills from a single clean demonstration, and show robust replay in the real world despite interruptions and perturbations. We will now describe each component in detail.

\vspace{-0.02in}
\subsection{Learning Low-level Skills}
\vspace{-0.02in}
Walking and manipulation using legs include drastically different joint angle behaviors. Hence, for training stability, we decouple the problem into locomotion policy $\pi_l$ to walk, and a policy to manipulate with legs $\pi_m$. Locomotion skill deals with movements, i.e. both walking and climbing, while manipulation involves using one leg to interact while using others to balance. Inspired by RMA~\cite{kumar2021rma}, we want to learn adaptive policies which use an online estimate of environment to adapt their behaviors appropriately. Such polices can walk in the real world under a diverse set of conditions.  Concretely, each policy takes the current state $o_t$ (roll, pitch, base angular velocities, velocity commands, joint positions and velocities, base velocity, feet contact indicators), the extrinsics vector $z_t$ (which is an estimate of the environment parameters), and additional task specific inputs. Then each policy outputs the target joint angles $a_t\in\mathbb{R}^{12}$ at 50Hz. We train these policies in simulation using model free reinforcement learning~\cite{Schulman2017-ky}. We will now describe the RL training setup which includes the additional task specific inputs, as well as the task specific reward terms which are used in addition to the following reward terms shared between both the policies to encourage smooth and energy efficient motions:  
\begin{itemize}[noitemsep,topsep=0pt]
    \item Joint Acceleration: $-||\ddot q_t||_2$
    \item Action Rate: $-||a_t - a_{t-1}||_2$
    \item Hip Position: $-||q_{hip}||_2$
    \item Work: $-|\tau^T_t \cdot \dot q_t|$
    \item Z Velocity: $-||v_z||_2$
\end{itemize}
where $v_z$ is vertical linear velocity, $\dot q_t, \ddot q_t, \tau_t$ are the joint velocities, accelerations and torques, and $q_{hip}$ is the position of the hip motor. These reward terms are respectively weighted with 2.5e-7, 5e-3, 0.1, 3e-3, 1.

\mypara{Locomotion Policy}
The goal of the locomotion policy $\pi_l$ is to follow a target command velocity ($v_x^{cmd}$, $v_y^{cmd}$, $\dot\omega_{yaw}^{cmd}$). The terrain in simulation is half fractal terrains, and the other half is slopes of different incline $\theta_{slope}$. The robot is initialized in the plane and its task is to walk closer to the wall and then climb it. Once it climbs the wall stably, it is commanded to de-climb the wall. To avoid local minima, we introduce \textit{Terrain Curriculum (TC)} where the incline of the slope is gradually increased during the course of the training all the way to a vertical wall. 

In addition to $o_t$ and $z_t$ described above, the robot observes, $d_w$, the closest distance from robot's base to wall edge (where the terrain starts to incline) clipped in $[0.25, 0.65]$m. We add the following task specific rewards:
\begin{itemize}[noitemsep,topsep=0pt]
    \item Linear Velocity Tracking: $exp(-5 ||v_{xy} - v^{cmd}_{xy}||_2)$
    \item Angular Velocity Tracking: $exp(-5 ||\dot\omega_{yaw} - \dot\omega_{yaw}^{cmd}||_2)$
    \item Goal Reaching: $1$ if $d_l > d_{thresh}\ \&\ d_r > d_{thresh}$ else $0$
    \item Feet Air : $\sum_{f=0}^4\left(\mathbf{t}_{a i r, f}-0.5\right)$
    \item Slope State Regularization: $-||q_t - q_{manip}||_2$ if robot is holding on the wall
\end{itemize}

where $v_{xy}=[v_x, v_y]^T$ are the base linear velocities, $v^{cmd}_{xy}=[v^{cmd}_x, v^{cmd}_y]^T$ is the commanded velocity, $\dot\omega_{yaw}$ is the yaw rate and $\dot\omega_{yaw}^{cmd}$ is the commanded yaw velocity, $\mathbf{t}_{a i r, f}$ is the time in air for foot $f$. $q_t$ is the joint position. $d_l$ and $d_r$ are the distances from the front left and right foot to the wall. 
$d_{thresh} =  0.7$ and $q_{manip}$ is the resting pose on the wall from which $\pi_m$ will get initialized.The weight values for each of the reward terms are: 1.5, 0.5, 1.5, 1.0 and 0.3 respectively.

\mypara{Manipulation Policy}
\label{subsec:manip}
Manipulation policy's goal is to follow a desired end-effector position $p_{foot}^{cmd}(t) = [p_x(t), p_y(t), p_z(t)]^T$ such that the foot can track any arbitrary pre-planned trajectory, even while resting on an inclined wall.
For simplicity, we formulate all the trajectories as a sinusoidal trajectory in $[0, \pi]$ resulting in a behavior that first lifts the foot and then drop it. 
By specifying a start point $p_{foot}^{cmd}(0) = [p_x(0), p_y(0), p_z(0)]^T$ and an end point $p_{foot}^{cmd}(T) = [p_x(T), p_y(T), p_z(T)]^T$, we can interpolate a sinusoidal trajectory, parameterized by foot lift height $H$ and duration $T$.
\begin{gather}
    p_{foot}^{cmd}(t) = [s\Delta x+p_x(0), s\Delta y+p_y(0), H\sin (\pi s)] \\
    s = \frac{\min(t, T)}{T},\ \Delta x= p_x(T)-p_x(0),\ \Delta y = p_y(T)-p_y(0) \nonumber
\end{gather}
where t goes from $0$ to $T$.
Regardless of slope angle $\theta_{slope}$, we always want the foot lift direction to be perpendicular to the slope plane. So $p_{foot}^{cmd}(0)$ and $p_{foot}^{cmd}(T)$ are all defined in the coordination frame whose origin is on the wall edge and xy axes are on the slope plane. 
Along with $o_t$ and $z_t$ described above, the policy additionally takes the timer variable $s$, foot lift height $H$, end point position viewed from robot's base frame $p_{foot_{base}}^{cmd}(T)$ as inputs.
We add following rewards:
\begin{itemize}[noitemsep,topsep=0pt]
    \item Trajectory Tracking: $exp(-10(p_{foot}^{cmd}(t) - p_{foot}(t))$
    \item Slope States Regularization: $-||q_t - q_{manip}||_2$
\end{itemize}
where $p_{foot}(t)$ is the foot position in the same frame mentioned above. The weights for the terms are 2.0, 0.5. The sampling ranges for low-level commands are in Tab. \ref{tab:cmd-ranges}.

\subsection{Simulation to Real Transfer}
We do not have access to the latent extrinsics $z_t$ in the real world to deploy the above policies. To resolve this, we follow~\cite{kumar2021rma} and learn to estimate $z_t$ from proprioception history. This is done by the adaptation module which we can train in simulation itself using supervised learning since we have access to ground truth values. However, we find that the adaptation module might not be able to faithfully reconstruct $z_t$, leading to an inferior overall performance. One way to recover the performance is to add a third phase by finetuning the motor policy with the estimate $\hat{z_t}$ instead of ground truth $z_t$, as in A-RMA~\cite{kumar2022adapting}. However, in this paper, we simplify the setup by reducing three phases into a joint single stage training which internally alternates between training a teacher policy with ground truth environment dynamics parameters ($z_t$) and an estimator of $z_t$ from proprioception history, inspired from Regularized Online Adaptation (ROA)~\cite{Fu2022-si}. Specifically, the privileged information encoder $\mu$ encodes information about the environment $e_t$ into a latent vector $z_t\in\mathbb{R}^{20}$ and then passes to the base policy ($\pi_l$ or $\pi_m$). The adaptation module $\phi$ trained using supervised learning by $z_t=\mu(e_t)$, while the privileged info encoder $\mu$ is encouraged to be close to $\phi(x_{t-10}, a_{t-11}, \cdots , x_{t-1}, a_{t-2})$. The ranges of environment dynamics randomization is same as ~\cite{kumar2021rma}.

However, unlike~\cite{Fu2022-si}, our robot needs to perform highly varying and agile tasks like climbing during which on-board sensing can be noisy. Hence, we need a way to estimate the values such as robot's velocity or foot contacts in an online manner itself. To facilitate this, we train an additional unified estimator $\omega$ to estimate unavailable robot states (such as velocity and foot position).
The estimated robot states are the base linear velocity $v_{base}=[v_x, v_y, v_z]^T$ for locomotion policy $\pi_l$, and manipulation foot position $p_{foot}=[x_{foot}, y_{foot}, z_{foot}]^T$ in robot's base frame. Similar to~\cite{Ji2022-lt}, we learn an estimator of $v_{base}$ and $p_{foot}$ from proprioception history in simulation. We call this Unified State Estimator (USE). We show that USE, ROA, and TC, all are integral to achieving good performance.

\begin{table}
        \resizebox{\linewidth}{!}{
        \begin{tabular}{c|c|c|c}
            \toprule
            Policy & Command Vars & Training Ranges & Test Ranges \\ \hline
            \multirow{3}{*}{Loco} & 
            $v_x^{\text{cmd}}\ (m/s)$ &  [-1.0, 1.0] & [-1.5, 1.5] \\
            &$v_y^{\text{cmd}}\ (m/s)$ &  [-0.6, 0.6] & [-1.0, 1.0] \\
            &$\dot\omega_{yaw}^{\text{cmd}}\ (rad/s)$ & [-0.5, 0.5] & [-0.6, 0.6] \\
            \midrule\multirow{4}{*}{Manip} & 
            $\Delta x\ (m)$ &  [-0.1, 0.3] & [-0.15, 0.3] \\
            &$\Delta y\ (m)$ &  [-0.1, 0.15] & [-0.2, 0.2] \\
            &$H\ (m)$ &  [0.06, 0.2] & [0.05, 0.22] \\
            &$T\ (s)$ &  [0.5, 2.0] & [0.3, 3.0] \\
            \midrule
            Common&$\theta_{slope}\ (rad)$ &  [0, 1.57] & [0, 1.57] \\
            \bottomrule
        \end{tabular}}
\caption{\small Uniform sampling ranges of low-level commands}    
    \vspace{-0.15in}
    \label{tab:cmd-ranges}
\end{table}

\subsection{Composing Skills into High-Level Behaviors}
Finally, we combine the above learned skills into long-range behaviors from only one demonstration. Note this single long-range demonstration is only in the high-level action space where human chooses what skill to follow and when, the low-level control is taken care of by the skills ($\pi_l$, $\pi_m$) learned above. We distill this single demonstration into a behavior tree~\cite{ogren2018bt} to learn to robustly complete the task.

\mypara{Behavior Tree Execution}
We assume our tasks are sequential and formulate our tree structure shown in Fig. \ref{fig:method}. The behavior tree executes from the starting root node, sequentially checking the children's condition node to see which ones have succeeded, and executing the one yet to be completed. If all children of the root node returns success, then the entire task will be completed.

\mypara{Behavior Tree Learning} The human gives one demonstration of a long-term task. During the demonstration, we record the human commands and the corresponding visual representation vectors from the second to last layer of ResNet18 \cite{resnet, shah2021rrl} $v^r\in\mathbb{R}^{512}$.
From the demonstrations, we create a tree whose number of children is equal to the number of skills executed $n_s$. The goal of child node $i \in [0, n_s-1]$ is to execute the $i^{th}$ high-level command until the corresponding condition node $Task\_i_{cond}$ returns true. 
The criteria for deciding $Task\_i_{cond}$ is a visual representation score in the form of cosine distance between current representation vector $v^r_t$ and that of expert's $D_r[i]$ when its subskill is complete, $score_i = 1 - \frac{v^r_t \cdot D_r[i]}{||v^r_t||_2 \cdot ||D_r[i]||_2}$. 
$Task\_i_{cond} = 
\begin{cases} 
True, & score_i < 0.16 \\ 
False, & Otherwise\end{cases}$

\mypara{Robustness via Task Precondition}
Every task has a precondition $Task\_i_{pre} = f(C[i], o_t, p_t, c_t)$, which is a function of the current robot states. 
If $Task\_i_{pre}==True$ is not satisfied, we rewind to the previous task recursively until we find a proper task that satisfy its own precondition.
The tree structure introduced above is robust to interruptions or perturbations by design. For example, if the robot slips right before pressing the button high up on the wall, 
then the precondition variable $Task\_i_{pre}$ of climbing on the wall will be $False$, and that will allow the behavior tree to backup, and re-execute the nodes which are incomplete.

\begin{figure}[t]
    \centering
    \begin{subfigure}[t]{0.47\linewidth}
    \centering
    \includegraphics[width=\linewidth]{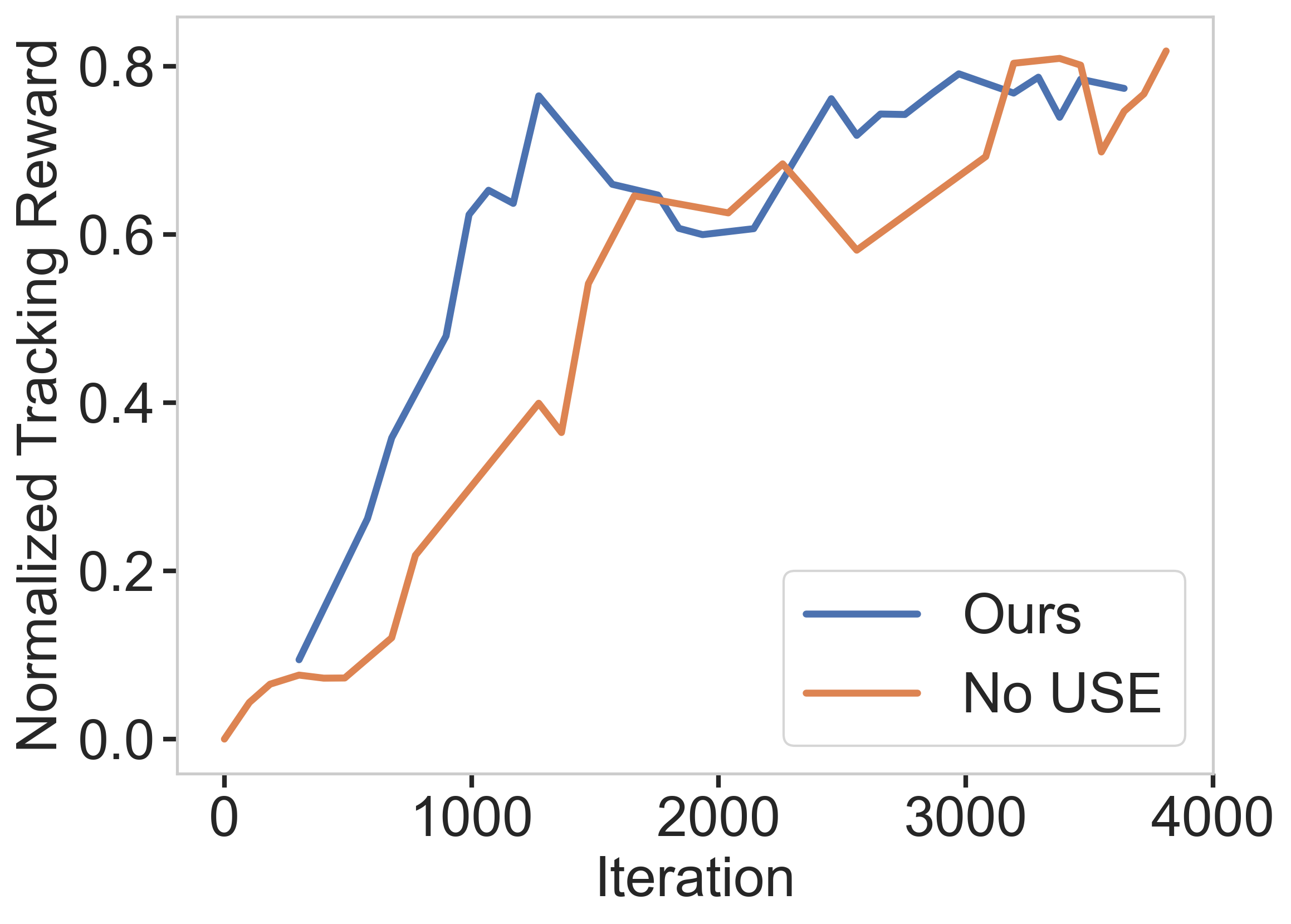}
    \vspace{-0.25in}
    \caption{\small Tracking performance}
    \end{subfigure}
    \quad
    \begin{subfigure}[t]{0.47\linewidth}
    \centering
    \includegraphics[width=\linewidth]{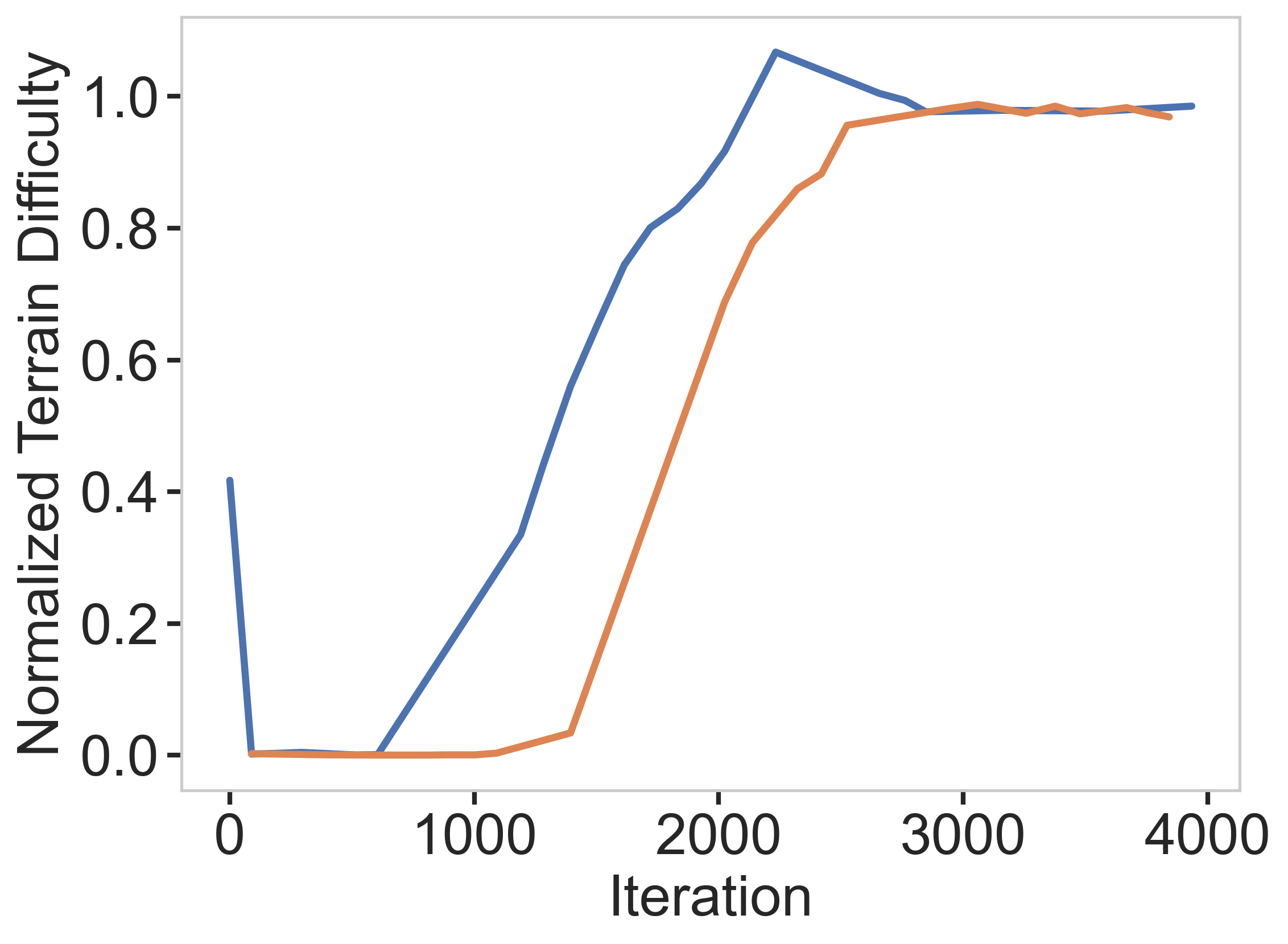}
    \vspace{-0.25in}
    \caption{\small Terrain difficulty}
    \end{subfigure}
    \vspace{-0.05in}
    \caption{\small USE helps the locomotion policy to achieve better performance and also boosts the training speed. The Normalized Terrain Difficulty $ntd$ indicates the maximum slope angle that the policy can climb $\theta_{slope}$. $ntd=1$ means $\theta_{slope}$ is uniformly sampled from the range in Tab. \ref{tab:cmd-ranges}. USE accelerates the policy learning, showing faster convergence to final performance for  linear and angular velocity tracking. USE also helps the policy to advance to more challenging terrains with a larger Slope Angle $\theta_{slope}$ faster.}
    \label{fig:ablation}
    \vspace{-0.18in}
\end{figure}

\section{Results, Setup and Analysis}
We quantitatively analyze a) the benefit of USE, ROA, and TC in simulation, b) compare our method with baselines on long-range tasks with interruptions, and c) show the robustness of our method on a very long-range task in the real world to show how our method enables the quadruped to expand its reach in real-world environments.

\mypara{Hardware and Simulation Details}
We use the Unitree Go1 quadrupedal robot and one forward-facing Realsense D435i camera to estimate distance to the wall and another side-facing one to detect an AprilTag\cite{Olson2011-qh} for button location. We use IsaacGym~\cite{makoviychuk2021isaac} as our simulation platform. 

\begin{figure*}[t]
    \centering
    \includegraphics[width=1.0\linewidth]{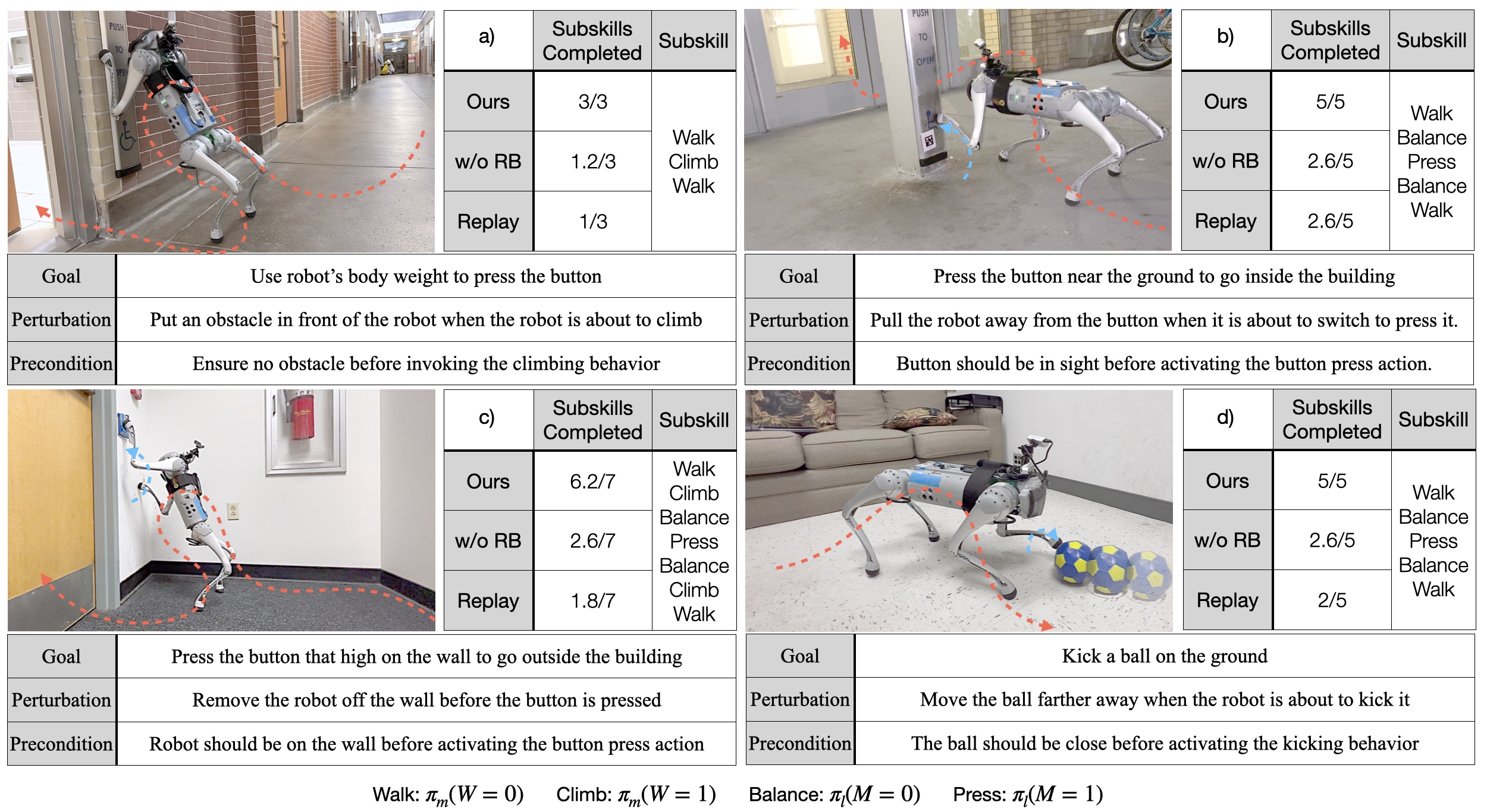}
    \vspace{-0.17in}
    \caption{\small Our method outperforms baseline methods and achieves the highest subskills completion rate. The red dashed line indicates the trajectory of the robot base. The blue line indicates the manipulation foot trajectory when $\pi_m$ is in use. We list all the subskills at the bottom and also list the sequence of skills needed for each long-range task next to it. Subskills Completed measures the fraction of selector nodes marked as a success. The data is averaged during 5 real-world trials for each task.}
    \label{fig:tasks}
    \vspace{-0.1in}
\end{figure*}

\begin{table}[t]
    \centering
    \resizebox{\columnwidth}{!}{
    \begin{tabular}{c|c|c|c|c}
        \toprule
             & Vel Error$\downarrow$ & Traj Error$\downarrow$ 
             & \makecell{$z_t$ loss $\downarrow$ \\ Loco} & \makecell{$z_t$ loss $\downarrow$ \\ Manip} \\ \hline
        Ours & $\textbf{29} \pm 11$ & $\textbf{3.1} \pm 3.0 $ 
        & $8.6\pm6.3$ &$\textbf{6.5}\pm6.0$\\
        w/o USE & $32 \pm 19$ & $3.7 \pm 3.9$ 
        & $\textbf{7.2}\pm3.6$ & $6.8\pm7.1$ \\ 
        RMA (\cite{kumar2021rma}) & $39 \pm 30$ & $4.7 \pm 6.1$ 
        & $92\pm33$ & $53\pm27$\\
        w/o TC & $58 \pm 17$ & - & $9.3\pm5.5$& -\\
        DR  & $41\pm31$ & $5.8\pm8.8$ & - & - \\ \hline
        Ours (priv) & $24 \pm 9.0$ & $3.2 \pm 3.3$ & - & - \\ 
        RMA (priv) & $23 \pm 6.8$& $3.1 \pm 3.4$ & - & - \\
        \bottomrule
    \end{tabular}}
    \caption{\small Low-level commands tracking performance in simulation ($1e-2$). Our method outperforms all baselines in tracking error. The $z_t$ loss is defined as $||\mu(e_t)-\phi(x_{t-10}, a_{t-11}, \cdots , x_{t-1}, a_{t-2})||_2$. We collect data equivalent to 10 hours real-world time. In Domain Rand, the robot does not observe environment extrinsics $z_t$.}
    \label{tab:ablation}
    \vspace{-0.15in}
\end{table}

\subsection{Simulation Experiments}

We show the quantitative importance of the Unified State Estimator (USE), Regularized Online Adaptation (ROA), and Terrain Curriculum (TC) in learning low-level skill execution in simulation across the following metrics: a) velocity tracking error, b) trajectory tracking error, c) $z$ regression error. We also compare to \textit{DR} (Domain Randomization), and Policies with ground truth extrinsics $z_t$ (\textit{Ours-priv} and \textit{RMA-priv}). We add external pushes during evaluation (Tab~\ref{tab:ablation}).

\mypara{Regularized Online Adaptation} The regularization does not degrade the expert's performance in ROA, while still showing a significant improvement in the student policy (about $\textbf{30}\%$ for $\pi_l$ and $\pi_m$) compared to RMA~\cite{kumar2021rma}, achieving an order of magnitude improvement in regressing to $z_t$.

\mypara{Unified State Estimator} To demonstrate the effectiveness of using USE in locomotion policy, we compare how USE improves policy learning during training as shown in Fig. \ref{fig:ablation}. With USE the policy reaches similar performance 10\% faster than the baseline method which does not employ USE.

\mypara{Terrain Curriculum} Without TC, the velocity error is $\textbf{107}\%$ larger than our method since the policy simply learns to walk until approaching the wall and stop. With TC, the robot can learn to climb the relatively challenging vertical wall by first learning to climb a slope with a lesser incline.

\subsection{Evaluating Robust Skill Composition in Real World}
We first compare our method with baselines on long-range tasks with interruptions to test the robustness in the real world. We define 4 setups that require the use of legs for manipulation (Fig~\ref{fig:tasks}) with the number of subskills $n_s \in \{3, 5, 7\}$. For each of these settings, we learn the behavior tree from one clean demonstration and evaluate the performance of our method with the stated interruption. Since we only have one interruption per task, we simplify our experiment design and define only one precondition corresponding to the subskill affected by the interruption. This precondition can be used to determine if the behavior tree needs to rollback.  

Fig. \ref{fig:tasks} compares our method's performance to baselines:
\begin{itemize}
    \item w/o RB: Use behavior tree to specify high-level commands but without task precondition. This version runs the nodes in sequence, switching only when the current task is completed, but fails to roll back to redo a previously finished task in case of an interruption. 
    \item Replay: Record the expert's high-level commands with time stamps, and simply replays them for the same amount of time as in the demonstration.
\end{itemize}

\begin{figure*}[t!]
    \centering
    \includegraphics[width=1.0\linewidth]{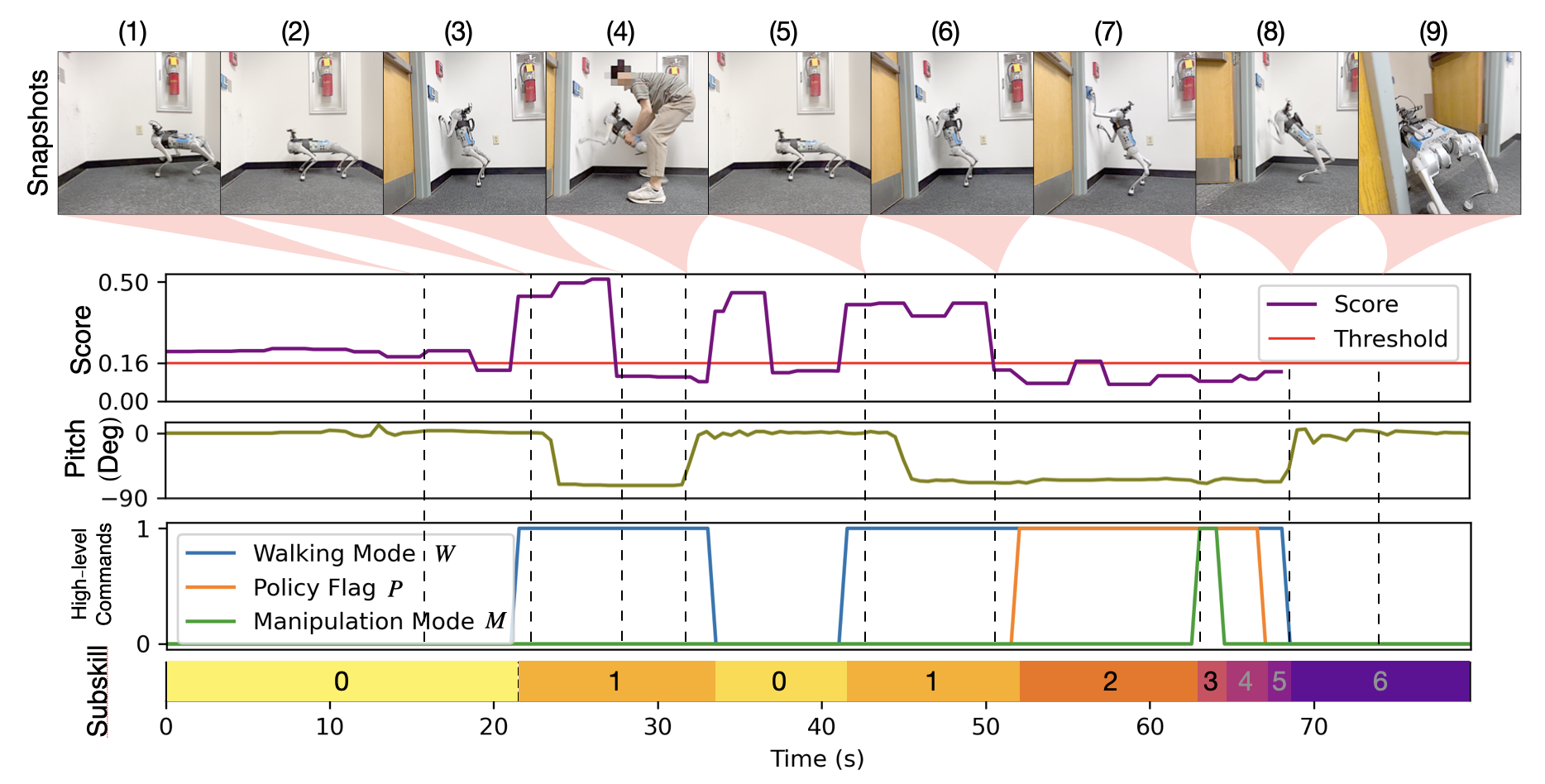}
    \vspace{-0.25in}
    \caption{\small Demonstration of a long horizon door-opening task with learned behavior tree from expert demonstration. The subskill tuples are specified in bottom left of Fig. \ref{fig:tasks}. The score is the cosine distance between the current visual representation vector and that of the expert's when the expert activates that subskill. Pitch is the pitch angle of the robot base. Subskill indicates which skill in the behavior tree is currently executing. Note that for subskill 6 there is no score because that is the last one.}
    \label{fig:exp-bt}
    \vspace{-0.12in}
\end{figure*}

We measure the subskill completion rate, as well as the overall task completion rate for each of the tasks across 5 trials (Fig.~\ref{fig:tasks}). We can see that our method achieves the highest subskills completion rate across trials. 
While w/o RB can complete subskills until the interruption, it is unable to complete the entire task. In a), w/o RB can sometimes climb the wall with the box obstacle under its body. In comparison, our method can switch back to walking mode and find another spot without an obstacle to complete the task. In b), w/o RB stretches its leg to the limit and manages to press the button if the robot is not moved too far away, while our method will switch back to walking when the button is too far.
In c), w/o RB will never climb the wall again when it is removed off the wall while our method falls back to climbing again and then completing the task.
Replay performs the worst since it uses no feedback on task completion. It naively executes the commands with the same timestamps as recorded during the expert demonstration.

\subsection{Analysis on a Long-Horizon Task in the Real World}
We now show a detailed analysis of how our framework works on a task that involves 7 subskills and 6 skill switches as shown in Fig. \ref{fig:exp-bt}. This task shows the expanded capability enabled by our low-level skills as well as the robustness of skill composition. The robot is able to press a wheelchair access button that is $0.95m$ above the ground and walk out of a door.
This task itself requires the robot to utilize two rear legs to raise the base (which is otherwise only 0.5m high), and then use its legs for interaction. It stretches the left front leg to reach the otherwise impossible-to-reach button and then presses it.
The button is located in a corner surrounded by walls on three sides. The space is about $2m$ long and $0.5m$ wide, which tight relative to the robot's dimensions ($0.5m$ long and $0.3m$ wide). The accurate low-level command tracking performance makes it precise enough to enable movement in this tight space.

In detail, the robot first needs to walk from an open field to this narrow corner as shown in Snapshot $(1)$. This is recorded by the behavior tree as a success after verifying that the visual representation score is smaller than the threshold. The robot then switches to climbing and climbs onto the wall ($(2)$, $(3)$). At this time, we manually perturb the robot and remove it off the wall and on the flat ground. Without task precondition, the behavior tree continues to try to press the button, although it first needs to climb the wall again. This leads to unpredictable and unsafe joint movements since it has never been trained in this scenario. However, with task precondition (\textit{ours}), the removal from the wall violates the precondition for button pressing, since the pitch angle of the robot exceeds -1. 
The behavior tree goes back to the previous task that satisfies its own precondition ($(4)$), and re-runs through the tree from that node. Consequently, the robot starts over to climb the wall ($(5)$, $(6)$), presses the button ($(7)$), and gets off the wall to go out the opened door ($(8)$, $(9)$). The system can handle multiple external perturbations, although, we only show 1 successful recovery here. 

\section{Conclusion}
In this paper, we present a framework for synthesizing low-level skills for quadrupedal robots that involve locomotion and manipulation using legs. We then compose these skills to achieve long-range complicated tasks that gives robots more capability to access human environments. Our framework is both agile and robust and aims to push the limits of robotic quadrupeds.
Our robot performs a set of useful real-world tasks including opening doors by pressing various types of buttons of different heights and interacting with moving objects. Rigorous evaluations are performed both in simulation as well as real world scenarios.
A limitation is that we currently decouple high-level decision making and low-level command tracking and making it fully end-to-end is an exciting future direction.
\clearpage
\balance
\section*{ACKNOWLEDGMENT}
This work was supported by NSF IIS-2024594 and DARPA Machine Common Sense.

{
\bibliographystyle{IEEEtran}
\bibliography{main}

}

\end{document}